\documentclass[runningheads]{llncs}
\usepackage{graphicx}
\usepackage{float}
\usepackage{graphicx} 
\usepackage{color} 
\usepackage[dvipsnames]{xcolor}
\definecolor{brickorange}{RGB}{193,74,9}
\definecolor{blue2}{RGB}{162,191,254}
\definecolor{red}{RGB}{255,0,0}
\definecolor{green}{RGB}{0,255,0}
\definecolor{green}{RGB}{51,132,51}
\definecolor{pink}{RGB}{249,187,195}
\definecolor{blue}{RGB}{192,247,249}
\definecolor{yellow}{RGB}{250,224,40}
\definecolor{red}{RGB}{222,91,76}
\definecolor{purple}{RGB}{222,156,218}
\definecolor{orange}{RGB}{245,136,24}
\usepackage{caption} 
\usepackage{subfigure} 
\usepackage{wrapfig} 
\usepackage{amsmath,amssymb,marvosym} 
\usepackage{lipsum} 
\usepackage{array} 
\usepackage{hyperref}
\newcommand{\etal}{\textit{et al.}}
\usepackage[colorinlistoftodos]{todonotes} 
\usepackage{algorithm} 
\usepackage[noend]{algpseudocode}

\usepackage{pgfplots}
\usepackage{environ}
\usepackage{epstopdf}
\makeatletter
\newsavebox{\measure@tikzpicture}
\NewEnviron{scaletikzpicturetowidth}[1]{%
  \def\tikz@width{#1}%
  \begin{lrbox}{\measure@tikzpicture}%
  \BODY
  \end{lrbox}%
  \pgfmathparse{#1/\wd\measure@tikzpicture}%
  \BODY
}
\makeatother
\usepackage{enumitem}
\usepackage{makecell}
\usepackage{url}
\usepackage{wasysym}
\usepackage{multirow}
\usepackage{tikz}
\usepackage{sidecap}

\newcommand{\repeatthanks}{\textsuperscript{\thefootnote}}
\usepackage{hyperref}
\hypersetup{
	colorlinks=true,
	linkcolor=blue,
	filecolor=magenta,      
	urlcolor=cyan,
}
\begin{document}
	\title{Shape-Aware Complementary-Task Learning for Multi-Organ Segmentation}
	%
	%
	\author{Fernando Navarro\thanks{The authors contributed equally to the work.}\inst{1}, Suprosanna Shit\repeatthanks\inst{1}, Ivan Ezhov\inst{1}, Johannes C. Paetzold\inst{1}, Andrei Gafita\inst{3}, Jan Peeken\inst{2}, Stephanie E. Combs \inst{2}, \and Bjoern H. Menze\inst{1}}
	
	\authorrunning{Fernando Navarro et al.}
	\titlerunning{Shape-Aware Complementary-Task Learning for Multi-Organ Segmentation}  
	
	%
	
	\institute{Department of Informatics, Technische Universit\"at M\"unchen, Germany\\
		\and
		Department of Radiotherapy, Klinikum rechts der Isar, Germany\\
		\and
		Department of Nuclear Medicine, Klinikum rechts der Isar, Germany \\
		\email{fernando.navarro@tum.de}}

	\maketitle              
	\begin{abstract}
		Multi-organ segmentation in whole-body computed tomography (CT) is a constant pre-processing step which finds its application in organ-specific image retrieval, radiotherapy planning, and interventional image analysis. We address this problem from an organ-specific shape-prior learning perspective. We introduce the idea of complementary-task learning to enforce shape-prior leveraging the existing target labels. We propose two complementary-tasks namely i) distance map regression and ii) contour map detection to explicitly encode the geometric properties of each organ. We evaluate the proposed solution on the public VISCERAL dataset containing  CT scans of multiple organs. We report a significant improvement of overall dice score from $0.8849$ to $0.9018$ due to the incorporation of complementary-task learning.
	\end{abstract}
	\section{Introduction}
	In representation learning, auxiliary-tasks are often designed to leverage \textit{free-of-cost} supervision which is derived from existing \textit{target labels}. The purpose of including auxiliary tasks is not only to learn a shared representation but also to learn efficiently by solving the common \textit{meta-objective}. A group of auxiliary-tasks driven by a common \textit{meta-objective} often have \textit{complementary objectives}. For example, to detect an orange, one can define two sub-tasks: i) learn only the shape of the orange, and, ii) learn only the color of the orange. Here, learning the shape and the color complement each other to learn the common \textit{meta-objective}: how an orange looks. In this context we leverage complementary-tasks learning by jointly optimizing the common \textit{meta-objective} of multiple-tasks.\\
	
	Solving multiple tasks simultaneously \cite{kendall2018multi} is known to improve each task's performance when compared to learning them independently. Mutual information exchange between multiple tasks such as detection and segmentation drive a neural network towards learning a generalized shared representation \cite{misra2016cross}. A more recent success in multi-task learning is mask-RCNN\cite{he2017mask} which benefited from the combined object detection and instance segmentation tasks. Uslu et al. \cite{uslu2018multi} shows that learning vessel interior, centerline and edges as a set of complementary-tasks improves junction detection in the retinal vasculature. There is a new body of research which aims to efficiently balance losses of different competing tasks \cite{chen2017gradnorm,kendall2018multi}. In contrast, our main objective is to leverage multiple complementary-tasks which shares a common \textit{meta-objective} and do not need additional annotated targets. Given that medical imaging problems such as multi-organ segmentation have to be solved with limited annotated data, complementary-task learning is a promising alternative solution.\\
	
	Multiple approaches have been proposed for multi-organ segmentation, which can be classified into registration-based and machine-learning based approaches. In registration-based methods, an atlas is registered to a test volume to obtain the segmentation map. \cite{wang2012multi,song2017progressive}. However this is a time-consuming method and its performance also suffers from inter-subject variability. Alternatively, the state-of-the-art methods for multi-organ segmentation are based on deep learning architectures such as fully convolutional neural networks FCN \cite{shelhamer2017}and U-Net \cite{ronneberger2015}. Deep learning-based methods can successfully segment large anatomical structures but are prone to miss small organs. Recently, Zhao \etal \cite{zhao2019knowledge} have explored the idea of combining both registration-based approaches and deep learning to segment small organs.\\ 
	
	However a contemporary study by Geirhos et al. \cite{geirhos2018imagenet} shows that convolutional neural networks are inherently biased towards texture information over the shape of an object. In human anatomy, all organs have a discriminative shape feature which has not been addressed at its full potential by previous approaches. Our hypothesis is that learning \textit{shape-prior} as a complementary-task improves the performance of the segmentation task. For learning the meta-objective of multi-organ representation, we define two complementary-tasks for shapes-prior learning: i) inferring geometric shape properties of an organ, and, ii) detecting the exterior contour of an organ. For the former, we propose the distance transform \cite{bischke2017multi} of shape as geometric shape properties. For the latter, we leverage the binary edge-map of each organ.\\
	
	\noindent
	In summary, our key contributions in this work are as follows:
	\begin{itemize}
		\renewcommand{\labelitemi}{$\bullet$}
		\item To the best of our knowledge, this is the first work for multi-organ segmentation leveraging complementary-task learning.
		\item We introduce two complementary-tasks in context of organ-specific shape-prior learning. We show that the inclusion of these complementary-tasks alongside the segmentation task improves its overall performance.
	\end{itemize}
	\begin{figure}[ht]
		\begin{center}
			\includegraphics[width=0.8\textwidth]{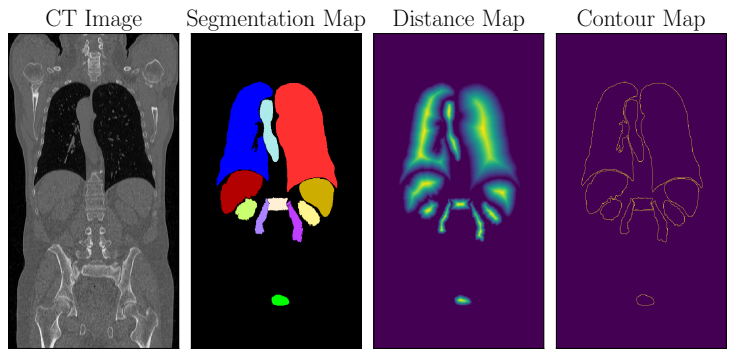}
		\end{center}
		\caption{\footnotesize{Target generation for complementary-tasks: \textit{Distance Map} is cumulative sum of the normalized distance transform of each organs' segmentation map. \textit{Contour Map} is the binary edge of each organ. Shape representation learning is jointly enforced in the distance and contour maps.}}
		\label{fig:overview}
	\end{figure}
	\section{Methodology}
	In  this  section,  we  present the proposed complementary-task learning for multi-organ segmentation and, subsequently describe the network architecture and the associated loss functions.\\

	In general, learning shape prior is itself a difficult task for a convolutional network due to the variety of shapes in different organs. We identify that the task of learning shape-priors can be broken down into multiple easier and quantifiable sub-task. Motivated by this fact, we propose two complementary-tasks to explicitly enforce shape and anatomical positional prior to the network.\\
	
	\noindent
	\textbf{Learning Distance-transform:} Euclidean distance transform of a shape is commonly used to find the inscribed circle having largest radius within an arbitrary shape. It maps a boundary regularity measure of a shape with respect to an interior point. Hence we argue that the distance transform regression learns geometric properties of shapes. In addition to that, we find that Gaussian heat-map regression is a common task in localizing anatomical landmarks \cite{payer2019integrating}. Parallels can be drawn between Gaussian heat-map regression and Euclidean distance map regression for soft-localization of organ-specific landmarks.
	\begin{figure}[ht]
		\begin{center}
			\includegraphics[width=1.0\textwidth]{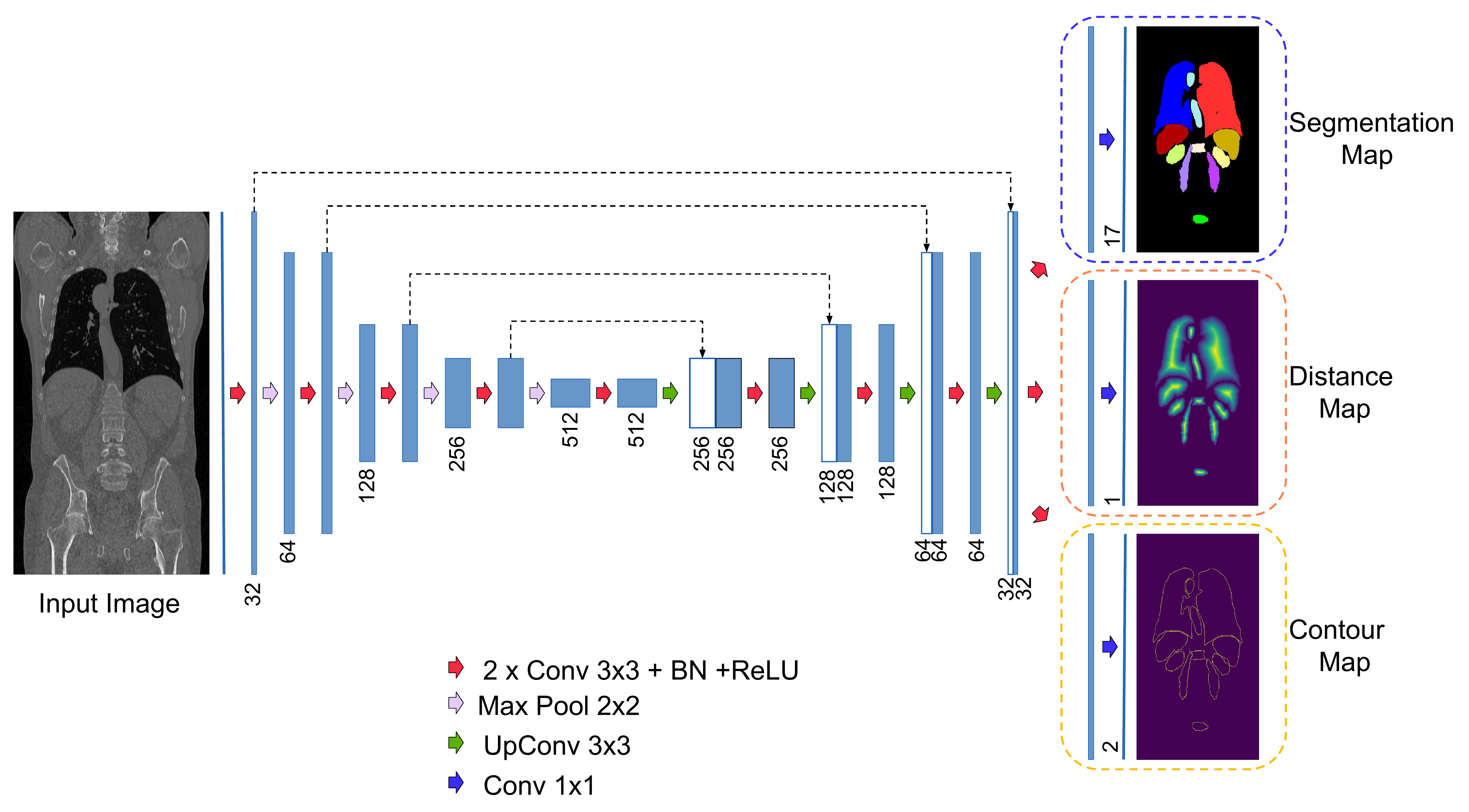}
		\end{center}
		\vspace{-0.5cm}
		\caption{\footnotesize{Network architecture for complementary-task learning. The input is a CT slice. The network consists of an encoding and decoding part with skip connections resembling the U-Net architecture with three branches diverging at the end of the last up-convolution. The outputs of the network are the segmentation map, the distance map, and the contour map.}}
		\label{fig:net}
	\end{figure}
	\\
	
	\noindent
	\textbf{Learning Organ Contour:} Detecting organ boundaries is the most challenging task for a segmentation network. Hence we propose to explicitly learn the organ contour as the second complementary-task alongside the distance map regression. The hypothesis is that the distance transform aids the network to accurately localize the organ from the learned anatomical prior whereas contour learning penalizes for boundary miss-classification, and, thus fine-tuning the organ shape.
	\begin{figure}[ht]
		\begin{center}
			\includegraphics[width=1.0\textwidth]{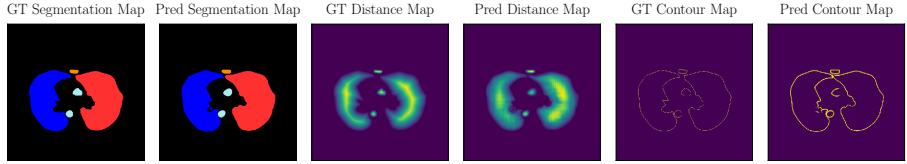}
		\end{center}
		\caption{\footnotesize{Qualitative comparison of an axial slice of a test sample between ground truth (GT) and the proposed model prediction (Pred). Our proposed model efficiently learn to i) segment the organs, ii) regress on distance map and iii) detect the organ contour map at the same time.}}
		\label{fig:pred}
	\end{figure}\\
	
	\noindent
	\textbf{Loss Function:} The loss function consist of three optimization objectives of the corresponding task: segmentation, distance regression and contour detection. Given the probability $p_{l}{(x)}$ of a pixel in location $x$ to belong to class $l$ and the ground truth by $g_{l}{(x)}$ the multi-organ segmentation loss $\mathcal{L}_{seg}$ and the contour map loss $\mathcal{L}_{contour}$ are defined as:
	\begin{equation}
	\mathcal{L}_{seg}, \mathcal{L}_{contour} = \underbrace{- \sum_{x}^{} {g_{l}{(x)} \log{p_{l}{(x)}}}}_\text{Cross-Entropy Loss} - \underbrace{{\frac{2 \sum_{x}^{} {p_{l}{(x)}} g_{l}{(x)}} {\sum_{x}^{} {p_{l}^{2}{(x)}} + \sum_{x}^{} {g_{l}^{2}{(x)}} } }}_\text{Dice Loss}
	\end{equation}
	where $l=\# \text{ of organs}+1$ in the multi-organ segmentation loss, and $l=2$ for the contour map. Dice loss is incorporated to handle the class-imbalance. For the distance map regression, the mean square error loss is optimized. Given the estimated distance map $p{(x)}$ of pixel $x$ and the ground truth value $g{(x)}$ the distance map loss function $\mathcal{L}_{dist}$ is:
	\begin{equation}
	\mathcal{L}_{dist} = - \frac{1}{n}\sum_{x}^{} { {\left({g{(x)} - {p{(x)}}}\right)}^{2}  } 
	\end{equation}
	where $n=\# \text{ of pixels}$ and the final loss $\mathcal{L}_{total} = \mathcal{L}_{seg} + \mathcal{L}_{contour}+ \mathcal{L}_{dist}$ is given by the summation of all losses. \\
	
	\noindent
	\textbf{Network Architecture:} The network architecture is inspired by encoding-decoding architectures with skip connections \cite{ronneberger2015,shelhamer2017}. A generalized shared feature representation is learned throughout the encoding blocks, the bottleneck, and a part of the decoding blocks of the network. Three different branches are added to the common feature representation to produce three output maps(c.f. Fig. \ref{fig:net}).\\
	
	\section{Experiments and Discussion}
	To validate the performance of the proposed approach, we perform the segmentation of 16 organs. The following four different experiments are performed:
	\begin{itemize}
		\renewcommand{\labelitemi}{$\bullet$}
		\item U-Net: baseline using only $\mathcal{L}_{seg}$.
		\item U-Net + distance: $\mathcal{L}_{seg} + \mathcal{L}_{dist}$.
		\item U-Net + contour: $\mathcal{L}_{seg} + \mathcal{L}_{contour}$.
		\item U-Net + distance,contour: $\mathcal{L}_{seg} + \mathcal{L}_{dist} + \mathcal{L}_{contour}$.
	\end{itemize}
	\textbf{Data-set:} The data-set used in the experiments consists of CT scans from the gold corpus and silver corpus in VISCERAL dataset \cite{jimenez2016cloud}. 74 CT scans from the silver corpus data were used for training, the annotations in this set are automatically labeled by fusing the results of multiple algorithms, yielding noisy labels. 23 CT scans from the gold corpus are used for testing, which contains manually annotated labels. Multi-organ segmentation imposes several challenges including the different fields of view (whole-body, thorax), modalities (contrast and non-contrast), severe class-imbalance across the target classes due to the diversity of organ's size and shapes and the capability to generalize when trained with noisy labels. 
	
	\begin{SCtable}
		\begin{tabular}{ll}
			\textbf{Model} & \textbf{Dice} \\
			\noalign{\hrule height 0.8pt}
			U-Net          & 0.8849 $\pm$ 0.120      \\ \hline
			U-Net + distance  & 0.8868 $\pm$ 0.116 \\ \hline
			U-Net + contour   & 0.8791 $\pm$ 0.118     \\ \hline
			U-Net + distance,contour & \textbf{0.9018 $\pm$  0.116} \\ \hline
		\end{tabular}
		\caption{\footnotesize Quantitative results: mean and standard deviation of the global dice scores shows that complementary-task learning achieves the best result.}
		\label{tab:results}
	\end{SCtable}
	\begin{figure}[ht]
		\begin{center}
			\includegraphics[width=1.0\textwidth]{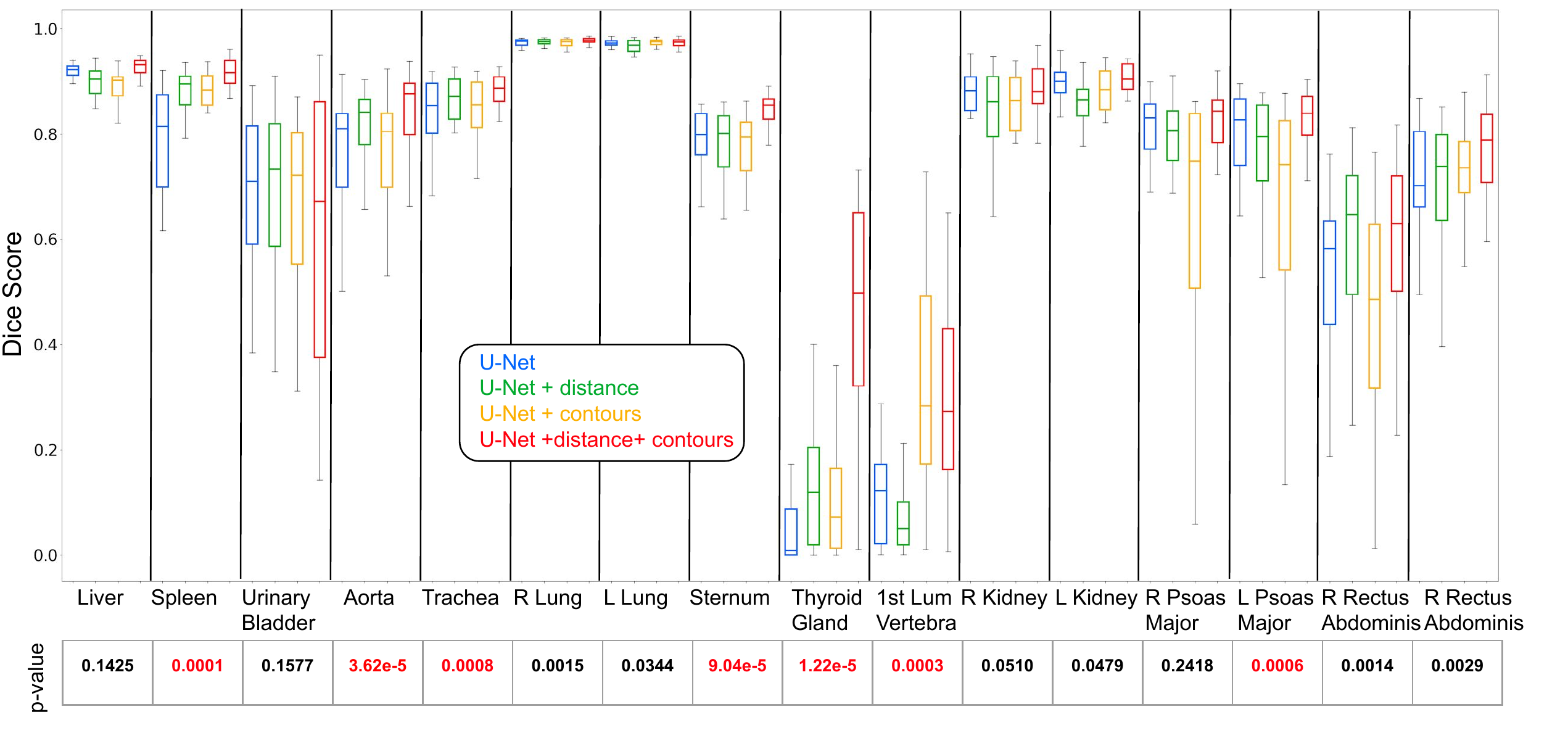}
		\end{center}
		\vspace{-0.2cm}
		\caption{\footnotesize{Box plots of different organs show consistent improvement of the dice score using our proposed model (U-Net+distance+contour) over the baseline model(U-net). We also report the \textit{p-value} obtained using Wilcoxon signed-rank statistical test between our proposed model and the baseline. The statistically significant \textit{p-values}$<0.001$ are shown in red.}}
		\label{fig:boxplot}
	\end{figure}
	All models were trained with Adam optimizer with a decaying learning rate initialized at 0.001, mini batch size of 4. Training was continued till
	validation loss converged. All the experiments were conducted on an NVIDIA Quadro P6000 GPU with 24GB vRAM. The code was developed in TensorFlow.
	
	\vspace{-0.1cm}
	To evaluate the overall accuracy of the proposed segmentation approach, we report the average dice score between the ground truth and the predicted segmentation map. Organ-specific dice scores are reported using box plots. We also report the Wilcoxon signed-rank test to find the statistical significance of the results.
	\vspace{-0.1cm}
	\subsubsection{Quantitative Results:} We observe from Table \ref{tab:results} that adding distance task improves the overall dice while adding contour task alone does not show any improvement. We attribute this behavior to the fact that the training data-set contains noisy labels and therefore the external part of the organs are not exactly represented. On the other hand, adding both tasks drives the segmentation network to better generalization resulting in an improvement of $2\%$ compared to the baseline. From the box plots in Fig \ref{fig:boxplot} we observe that dice score for big organs does not improve significantly compared to the baseline. In contrast, complementary-task learning provides statistically significant improvement for small organs such as the spleen, thyroid gland, and trachea.
	\begin{figure}[ht]
		\includegraphics[width=1.0\textwidth]{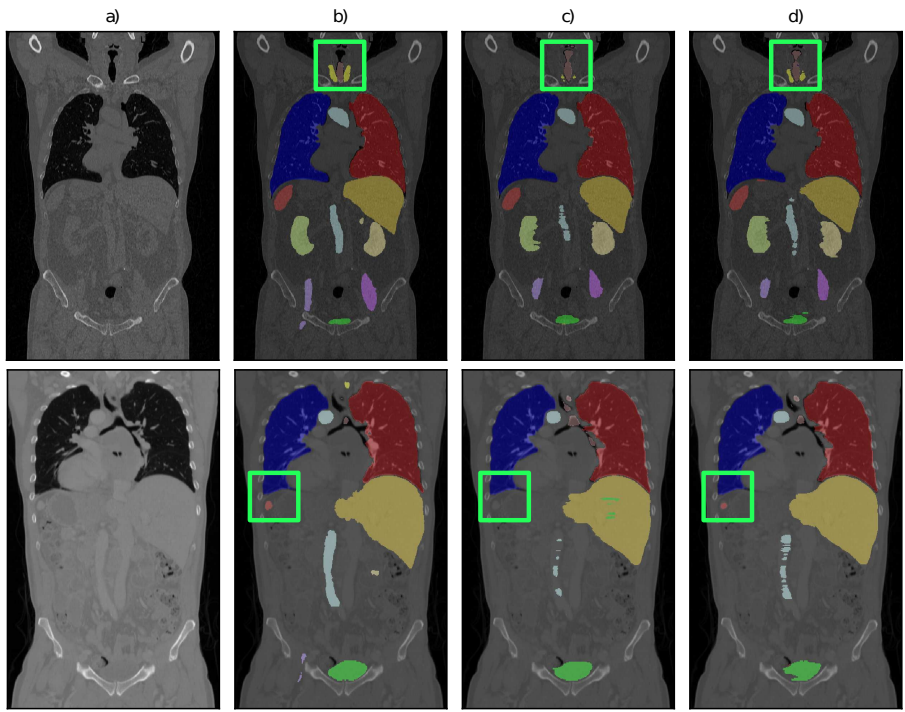}
		\caption[Qualitative results]{\footnotesize{Qualitative results: Each image shows the mid-slice in the coronal view. a) the input CT slice, b) ground truth segmentation, c) prediction form U-Net baseline, d) complementary-task learning. ROIs are indicated by green box highlighting regions where complementary-tasks learning improves the segmentation.}}
		\label{fig:results}
		\vspace{-0.2cm}
	\end{figure}
	\subsubsection{Qualitative Results:} Fig \ref{fig:results} shows a qualitative comparison between the U-Net baseline and the proposed approach. We highlight ROIs with a green
	box, to show regions where complementary-task learning improves the segmentation. We can visually asses that particularly the small organs results in better segmentation. Additionally, Fig \ref{fig:pred} compares the predicted distance map and the contour map to the ground truth. This confirms that the network is also able to solve the complementary-tasks.
	\vspace{-0.1cm}
	\section{Conclusions}
	In this work, we have proposed complementary-task learning to provide a novel and alternative solution to the challenging task of multi-organ segmentation. We have validated our method in a public benchmark data set which shows consistent improvement in dice score, especially for the small organs. In medical image segmentation where large data sets are scarce and corresponding dense annotation is expensive, designing complementary-task by leveraging existing target label could be beneficial to learn a generalized representation.\\ 
	\newline
	\noindent\textbf{Acknowledgements}
	\newline\\
	Fernando Navarro gratefully acknowledge the Deutsche Forschungsgemeinschaft (DFG, German Research Foundation) - GRK for the financial support. Suprosanna Shit and Ivan Ezhov are supported by the Translational Brain Imaging Training Network  (TRABIT) under the European Union’s ‘Horizon 2020’ research \& innovation programme (Grant agreement ID: 765148). The authors gratefully acknowledge the support of NVIDIA Corporation with the donation of the Titan Xp GPU used for this research.
	
	%
	%
	%
	\bibliographystyle{splncs04}
	\bibliography{mybibliography}

\begin{thebibliography}{10}
\providecommand{\url}[1]{\texttt{#1}}
\providecommand{\urlprefix}{URL }
\providecommand{\doi}[1]{https://doi.org/#1}

\bibitem{bischke2017multi}
Bischke, B., et~al.: Multi-task learning for segmentation of building
  footprints with deep neural networks. arXiv preprint arXiv:1709.05932  (2017)

\bibitem{chen2017gradnorm}
Chen, Z., et~al.: Gradnorm: Gradient normalization for adaptive loss balancing
  in deep multitask networks. In: Proceedings of the ICML (2018)

\bibitem{geirhos2018imagenet}
Geirhos, R., et~al.: Imagenet-trained {CNN}s are biased towards texture;
  increasing shape bias improves accuracy and robustness. In: Proceedings of
  the ICLR (2019)

\bibitem{he2017mask}
He, K., et~al.: Mask r-cnn. In: Proceedings of the IEEE CVPR. pp. 2961--2969
  (2017)

\bibitem{kendall2018multi}
Kendall, A., et~al.: Multi-task learning using uncertainty to weigh losses for
  scene geometry and semantics. In: Proceedings of the IEEE CVPR. pp.
  7482--7491 (2018)

\bibitem{misra2016cross}
Misra, I., et~al.: Cross-stitch networks for multi-task learning. In:
  Proceedings of the IEEE CVPR. pp. 3994--4003 (2016)

\bibitem{payer2019integrating}
Payer, C., et~al.: Integrating spatial configuration into heatmap regression
  based cnns for landmark localization. Medical Image Analysis  \textbf{54},
  207--219 (2019)

\bibitem{ronneberger2015}
Ronneberger, O., et~al.: U-net: Convolutional networks for biomedical image
  segmentation. In: Proceedings of the MICCAI (2015)

\bibitem{shelhamer2017}
Shelhamer, E., et~al.: Fully convolutional networks for semantic segmentation.
  IEEE TPAMI.  (2017)

\bibitem{song2017progressive}
Song, Y., et~al.: Progressive multi-atlas label fusion by dictionary evolution.
  Medical Image Analysis  (2017)

\bibitem{jimenez2016cloud}
Jimenez-del Toro, O., et~al.: Cloud-based evaluation of anatomical structure
  segmentation and landmark detection algorithms: {VISCERAL} anatomy
  benchmarks. IEEE TMI  \textbf{35}(11),  2459--2475 (2016)

\bibitem{uslu2018multi}
Uslu, F., Bharath, A.A.: A multi-task network to detect junctions in retinal
  vasculature. In: Proceedings of the MICCAI. pp. 92--100. Springer (2018)

\bibitem{wang2012multi}
Wang, H., et~al.: Multi-atlas segmentation with joint label fusion. IEEE TMI
  (2012)

\bibitem{zhao2019knowledge}
Zhao, Y., et~al.: Knowledge-aided convolutional neural network for small organ
  segmentation. IEEE JBHI  (2019)

\end{thebibliography}
\end{document}